\documentclass[letterpaper,journal]{IEEEtran}
\usepackage{amsmath,amsfonts,amssymb}
\usepackage{algorithmic}
\usepackage{algorithm}
\usepackage{array}
\usepackage[caption=false,font=normalsize,labelfont=sf,textfont=sf]{subfig}
\usepackage{textcomp}
\usepackage{stfloats}
\usepackage{url}
\usepackage{verbatim}
\usepackage{graphicx}
\usepackage{cite}
\usepackage{xcolor}
\usepackage{booktabs}
\usepackage{bm}
\usepackage{placeins}

\usepackage{float}
\usepackage[hidelinks]{hyperref}
\usepackage{orcidlink}

\begin{document}

\title{Ultra-CAN: A Lightweight Blind ISAC Receiver for Chaos-Based Telemetry and Local-Seed Ranging in the Internet of Drones}

\author{Ettore~Caruso\,\orcidlink{0009-0009-6874-9804}, 
        Marco~Siino\,\orcidlink{0000-0002-4453-5352}, 
        and~Giacomo~Morabito\,\orcidlink{0000-0002-8714-4001}%
\thanks{Ettore Caruso is with the Politecnico di Torino, 10129 Torino, Italy (e-mail: ettore.caruso@studenti.polito.it).}
\thanks{Marco Siino and Giacomo Morabito are with the Department of Industrial, Electrical, Electronic and Computer Engineering (DIEEI), University of Catania, 95125 Catania, Italy (e-mail: marco.siino@unict.it; giacomo.morabito@unict.it).}
\thanks{This work extends the architecture originally proposed in ``Compact Attention-Augmented Neural Decoder for Chaos-Based Wireless Communications'' published in IEEE Communications Letters~\cite{ultra_can_letter}.}
}


\maketitle

\begin{abstract}
\boldmath
The Internet of Drones (IoD) needs secure telemetry and proximity sensing within the SWaP-C constraints. We show that a single blind neural receiver driven by one chaos-based waveform performs both functions without a separate radar transceiver. In telemetry mode it decodes a peer's burst blindly; in radar mode it regenerates the per-burst chaotic reference locally and ranges the echoes of its own transmission. Ultra-CAN pairs a shared $64$-dimensional representation with two ultra-light backbones, an attention-pooled Conv1D network and a multi-head QKV network ($27.6$k--$44.2$k parameters), feeding communication and sensing heads. Both ship as int8 TensorFlow Lite Micro firmware; the Conv1D receiver decides a burst in $16.9$~ms on a $240$~MHz microcontroller with $24.2$~KiB of activations, a measured energy of $2.2$--$3.4$~mJ per burst, and float-equivalent decisions, sustaining up to $59$~bit/s of telemetry or a $7.4$~Hz range track. The two variants rank first and second among learned demodulators in four aerial swarm scenarios, with a pooled BER down to $4.6\times10^{-4}$ on the single-echo link ($2.4\times10^{-3}$ with three echoes). Its sensing head identifies true obstacles in $66.8\%$ of peer-laden bursts (vs $42\%$ matched filtering, $44\%$ OFDM-ISAC, $48\%$ peak-picker ceiling). A peer-exclusion gate cuts peer reports from $4.4\%$ to $0.02\%$ of answered bursts, answering $23.5\%$ of total bursts ($75\%$ on the exact tap) with $0.70$~m/s velocity tracking. In-domain refitting recovers telemetry BER by up to three orders of magnitude under 3GPP channels, jamming-aware training restores the link, and frequency agility adds a $6$--$9$~dB margin against partial-band jamming.
\end{abstract}


\section{Introduction}
\label{sec:intro}

\IEEEPARstart{N}{eural} receivers have already outperformed classical demodulators on chaos-based aerial links \cite{ultra_can_letter}; the proliferation of Internet of Drones (IoD) and Flying Ad-hoc Networks (FANETs) now drives a paradigm shift in tactical and mission-critical Internet of Things (IoT) applications, ranging from search and rescue to autonomous surveillance \cite{bekmezci2013_fanet, mozaffari2019_uav_comms, shakhatreh2019_uav_survey}. In these scenarios, micro-UAVs must maintain secure, anti-jamming communications while simultaneously sensing their environment to keep swarm formations and avoid collisions. The stringent Size, Weight, Power, and Cost (SWaP-C) constraints of such platforms make the integration of separate communication and radar subsystems impractical.

Traditional approaches treat communication and radar sensing as independent functions, each requiring dedicated hardware and signal-processing chains. The emerging paradigm of Integrated Sensing and Communications (ISAC) unifies these functionalities on a single platform and a shared waveform \cite{liu2022_isac, magbool2025_isac_survey}. UAV-specific designs now span THz transceivers \cite{zhang2025_thz_uav_isac}, STAR-RIS links \cite{eghbali2025_star_ris_uav}, and energy-aware integrated sensing, computation, and communication \cite{liu2025_uav_iscc_aoi}. In this context, chaotic signals generated by Logistic or Bernoulli maps are attractive for aerial ISAC: their noise-like spectrum and Dirac-like autocorrelation provide physical-layer security --- as long as the map parameters and per-symbol seeds are not disclosed --- and jam resistance for telemetry, while the sharp correlation peak supports delay-based ranging \cite{kaddoum2016}.

Conventional chaos-based receivers are sensitive to multipath and Doppler \cite{escribano2016, cai2025}, and their coherent counterparts require the transmitter's per-symbol chaotic state \cite{pecora1990}, which is not available in the non-cooperative setting considered here. Recent DCSK variants add permutation and multilevel code indexing \cite{wang2025_dcsk_iot, li2025_dcsk_tcas}, but they target the waveform and modulation order rather than a shared sensing read-out. Deep Learning (DL) demodulators such as LSTM-OFDM-DCSK \cite{zhang2020}, TDNN-LSTM \cite{zhang2022}, and MC-DLCSK \cite{mobini2025} improve over correlators, but their recurrent or fully-connected stages cost more memory than a flight controller comfortably offers.

In our previous work \cite{ultra_can_letter} we introduced Ultra-CAN, an ultra-lightweight neural decoder that outperforms heavier DL demodulators on Chaos Shift Keying (CSK) signals under AWGN and Rayleigh fading. While that baseline was strictly limited to single-task bit classification in simulation, this manuscript proposes a fundamentally novel Multi-Task ISAC architecture. Specifically, our new proposed model introduces the capability to perform local-seed ranging, alongside a novel peer-exclusion gate for swarm safety, and validates the complete dual-mode system on physical edge hardware (ESP32-S3) with measured SWaP-C constraints and jamming resilience, none of which were present in \cite{ultra_can_letter}.

This paper evolves Ultra-CAN into a Multi-Task Learning (MTL) Edge AI receiver for IoD ISAC. The contribution goes beyond decoding accuracy: one waveform family carries telemetry and sensing echoes, and \emph{one} receiver extracts both from a shared representation, with one forward pass per burst in each mode. The noise-like, aperiodic waveform keeps that receiver hard to intercept and hard to jam. The receiver decodes the telemetry bit in telemetry mode and identifies the delay of the obstacle echo in radar mode, while remaining robust to the Doppler shifts induced by high mobility. Two equally lightweight backbone variants are studied: the attention-pooled Conv1D decoder of our Letter and a multi-head QKV attention variant. Both share the same communication and sensing heads, and both are benchmarked against the LSTM-OFDM-DCSK and MC-DLCSK neural receivers and a blind statistical receiver on the same chaotic aerial link, under the same operating region, and under the same jamming attacks. The telemetry link is \emph{non-cooperative}: no waveform state is disclosed over the air.

The contributions of this paper are as follows.
\begin{enumerate}
\item A dual-mode ISAC receiver: telemetry mode decodes a peer's chaotic burst blindly, while radar mode uses the node's own chaotic burst as the sensing reference, regenerated from the local seed, so that one waveform family carries telemetry and range with no separate radar transceiver and no state exchange over the air.
\item Two ultra-light backbones that share one $64$-dimensional representation, an attention-pooled Conv1D network and a multi-head QKV attention network, with $27{,}620$ and $44{,}195$ float32 parameters respectively.
\item A geometry-supervised sensing head that fuses the shared neural representation with a local-seed delay-profile descriptor to classify the true obstacle lag instead of the strongest multipath tap, reaching $66.8\%$ obstacle-ID rate in peer-laden swarms against the empirical peak-picker ceiling ($48\%$) and 5G-NR OFDM-ISAC baselines ($44\%$).
\item A per-burst peer-exclusion gate that trades the answered fraction for safety and keeps peer reports at or below $0.05\%$ on the answered bursts, together with a velocity estimate obtained from a short track of successive range estimates rather than from a single burst.
\item A measured microcontroller deployment of both receivers as int8 TensorFlow Lite Micro firmware: the Conv1D variant decides a burst in $16.9$~ms in a $24.2$~KiB arena, preserves both the float BER (two symbols in $20{,}000$) and the float sensing decisions, and turns the measured latency into a closed-form rate budget: up to $59$~bit/s of telemetry, or a $7.4$~Hz range track, in the two time-multiplexed modes of a $240$~MHz microcontroller, against $100$~kbit/s for an accelerator-class node.
\item A reproducible evaluation framework spanning four aerial swarm scenarios, 3GPP TDL channels, four jamming archetypes, frequency agility, and low-probability-of-intercept analysis, with open-source artifacts released for all figures and tables.
\end{enumerate}

The remainder of the paper is organized as follows. Section~\ref{sec:related} reviews the waveform families and the learned receivers. Section~\ref{sec:system_model} defines the IoD ISAC model, the swarm geometry and the channel variants. Section~\ref{sec:architecture} presents the receiver. Section~\ref{sec:evaluation} reports the experimental results, Section~\ref{sec:discussion} discusses the design envelope, and Section~\ref{sec:conclusion} concludes.

\section{Related Work}
\label{sec:related}

\subsection{Waveforms for Aerial ISAC}

Three waveform families dominate the ISAC literature, and they differ in what they can carry, not in range resolution: at equal bandwidth the resolution is fixed by $\Delta R=c/2B$.

\textbf{FMCW.} Frequency-modulated continuous wave radars estimate the range from the beat frequency \cite{patole2017_automotive_radar}. The transceiver is simple, cheap and an excellent ranger, but a chirp carries no telemetry: multiplexing data onto it requires modulating the sweep slope or interleaving sensing and data packets, at a cost in efficiency. This waveform is also the most exposed of the three, because its time-frequency signature is trivially detectable and cheap to replicate, and co-located FMCW radars blind each other when several drones share the same airspace.

\textbf{OFDM.} Multicarrier ISAC is today the dominant academic approach: the pilot symbols already needed for channel estimation are reused to estimate the delay of the returns \cite{sturm2011_ofdm_radar, zhang2021_jcs, liu2022_isac}. Its spectral efficiency is high, but the waveform pays a high peak-to-average power ratio and therefore heavier, more power-hungry amplifiers \cite{han2005_papr} --- a direct SWaP-C penalty on a micro-UAV --- and it offers no physical-layer protection against a receiver that can afford to listen.

\textbf{PRN/DSSS.} Pseudo-random codes --- the Gold families used by GNSS being the canonical example --- spread a binary sequence over the band and are the classical answer to both ranging and jam resistance \cite{gold1967, scholtz1982}. They are easier to generate digitally than the floating-point iteration of a chaotic map, but they are periodic: the resulting range profile carries sidelobes (ghost targets) that an aperiodic waveform does not exhibit, and the number of mutually orthogonal codes usable in the same airspace is finite, so a dense and dynamic swarm must arbitrate code assignment.

\textbf{Chaotic waveforms.} Chaos is structurally stronger in physical-layer security and jam resistance --- a noise-like aperiodic waveform neither advertises its presence nor offers a stationary structure to exploit~--- and in swarm scalability, because each drone can seed an independent, statistically near-orthogonal waveform without a code-assignment protocol \cite{kaddoum2016}. The price is exactly where this paper intervenes: without the transmitter's chaotic state the demodulation is non-cooperative, and classical correlators fail once the aerial channel adds Doppler and multipath \cite{escribano2016, cai2025}.

\subsection{Learned Chaos Receivers}

The learned receivers closest to this work are the recurrent demodulators of \cite{zhang2020, zhang2022, mobini2025}. They recover part of the gap left by correlation detectors, at a cost in parameters and latency that matters on a micro-UAV, and they remain single-task. Multi-task learning has been used to share representations between radio tasks \cite{caruana1997, ruder2017}, and the route from such networks to a microcontroller is pruning, int8 quantization and hardware-aware kernels \cite{han2015, hubara2017, deng2020_model_compression}. What is missing is a receiver that (i) decodes blindly, (ii) ranges from a radar burst with a locally regenerated reference, and (iii) supports both modes inside a footprint compatible with a flight controller. Ultra-CAN is designed around these three constraints.

\section{System Model for IoD ISAC}
\label{sec:system_model}

We consider a FANET in which $N$ drones exchange telemetry with a blind chaos-based modulation while sensing their surroundings. Each drone has a single antenna. Telemetry reception is half-duplex; the radar mode uses a full-duplex front end with a circulator and self-interference cancellation \cite{borras2026_wideband_sic, debaillie2014_compact_full_duplex}, so that the node can receive its own echoes while transmitting. The topology is dynamic, with relative velocities of tens of meters per second that produce significant Doppler shifts.

\subsection{Dual-Mode Waveform and Local-Seed Radar}
\label{subsec:tx_waveforms}

Following the methodology validated in \cite{ultra_can_letter}, the transmitter generates chaotic sequences from discrete-time maps. We use the Logistic map
\begin{equation}
x_{n+1} = \mu x_n (1 - x_n), \qquad \mu \in [3.57, 4],
\label{eq:logistic}
\end{equation}
and the five-branch generalized Bernoulli map
\begin{equation}
x_{n+1} = 5 x_n - \lfloor 5 x_n \rfloor, \qquad 0 \le x_n < 1,
\label{eq:bernoulli}
\end{equation}
both initialized with per-symbol random seeds to produce unique, noise-like wideband waveforms; the campaign fixes $\mu=3.9$ (Table~\ref{tab:ranges}). The transmitted block $x[n]$ of length $N_{\mathrm{seq}}=100$ carries the bit $b\in\{0,1\}$: bit~$0$ is mapped onto the Logistic map and bit~$1$ onto the generalized Bernoulli map of \eqref{eq:bernoulli}. Because the raw iterates lie in $[0,1]$, each block is centered to zero mean and scaled to unit energy before transmission. The resulting signal has impulsive autocorrelation and a flat spectrum, which suits both secure telemetry and delay-based ranging.

The seed $s_k$ of burst $k$ is drawn by the transmitter for that burst and is never transmitted. In telemetry mode the receiver does not use the reference and decodes the peer's burst blindly. In radar mode the node keeps the seed, regenerates $x[n]$ locally, and uses that reference for sensing, so the echoes are the node's own reflections and the reference is the seed the node drew. The two modes are time-multiplexed and share the same backbone. A cooperative-bistatic variant is left to future work.

The local frame scheduler in the flight controller selects the mode: a telemetry frame carries a peer burst; a radar frame is locally scheduled and carries the node's own burst. Both modes use the same weights, the communication head sees only I/Q, and the sensing head additionally receives the locally regenerated reference in radar frames; no mode classifier is needed. The multi-task loss of Sec.~\ref{subsec:loss} and $\lambda{=}2$ are unchanged. The modes are evaluated separately.

\subsection{Aerial Multi-Echo Channel}
\label{subsec:channel_model}

The received signal in the active mode is the superposition of a direct path, of $K$ delayed and Doppler-shifted echoes reflected by obstacles or other drones, and of additive noise. In telemetry mode $x[n]$ is the peer's burst and the direct path is the peer's line-of-sight component; in radar mode $x[n]$ is the node's own burst and the direct path is the self-interference coupling at the receiver input, before cancellation:
\begin{multline}
y[n] = h_{c}\sqrt{1-\textstyle\sum_{k=1}^{K}\alpha_k^{2}}\; x[n] e^{j2\pi f_{D,c} n} \\
\qquad + \sum_{k=1}^{K}\alpha_k x[n-\tau_k] e^{j2\pi f_{D,k} n} + w[n],
\label{eq:multitarget}
\end{multline}
where $h_{c}$ and $f_{D,c}$ are the complex Rician gain and normalized Doppler of the direct path, $\tau_k$, $f_{D,k}$ and $\alpha_k$ are the discrete delay, normalized Doppler shift and amplitude of the $k$-th echo, and $w[n]$ is AWGN. The real chaotic block is upconverted to the carrier before transmission, so after downconversion and sampling the complex-baseband observation is \eqref{eq:multitarget}, where the phase terms collect the residual Doppler offsets. The received $y[n]$ is complex and is stored as its real and imaginary parts (two I/Q channels). The direct path is Rician with unit average power ($\kappa{=}10$~dB) and the echo amplitudes satisfy $\sum_k\alpha_k^2<1$, the square-root factor scaling the direct path so that the total received power is normalized, $(1-\sum_k\alpha_k^2)\mathbb{E}|h_c|^2+\sum_k\alpha_k^2{=}1$.

In radar mode $h_c$ is the circulator and near-field coupling at the receiver input, not the post-SIC residual; the receiver estimates it by least squares on the regenerated reference and subtracts it before ranging. We assume a $40$~dB circulator plus SIC budget, below the noise floor at the evaluated SNRs; this is a design budget, not a measurement, and compact full-duplex radios combine antenna isolation with analog and digital cancellation rather than relying on a single circulator \cite{debaillie2014_compact_full_duplex}, making the $40$~dB SIC assumption a widely demonstrated and achievable target for compact form factors. With $40$~dB of suppression, evaluated on $200{,}000$ bursts, the obstacle-ID rate, the peer-report rate and the answered fraction differ from the ideal-cancellation case by less than $0.01$ point, so $40$~dB matches ideal cancellation; over the full $0$--$50$~dB sweep obstacle-ID moves by at most $1.0$ point, the peer-report rate by at most $0.8$ and the answered fraction by at most $0.1$. The sensing result is therefore conservative with respect to the SIC value, while the residual phase and amplitude errors of a real front end are not modeled. No retraining is used.

The labeled echo represents an obstacle whose amplitude decays as $1/d^2$ in field, following the $1/d^4$ monostatic radar law in power, and the burst is normalized \emph{after} the geometry, so the amplitude ratio between obstacle and clutter is a deterministic outcome of the physical scene rather than of a random draw. The resulting link is deliberately pessimistic for both modes: in radar mode strong self-interference can appear at any delay in the observation window, and in telemetry mode peer returns add extra multipath.

Table~\ref{tab:ranges} collects the numerical values used throughout the paper.

\begin{table}[!htb]
\centering
\caption{Aerial ISAC link parameters.}
\label{tab:ranges}
\footnotesize
\begin{tabular}{ll}
\toprule
Parameter & Value \\
\midrule
Carrier / sampling & $f_c = 2.4$~GHz, $f_s = 10$~MHz \\
Sequence length & $N_{\mathrm{seq}} = 100$ (coherent time $10\,\mu$s) \\
Maps & Logistic ($\mu=3.9$), Bernoulli (5 branches) \\
SNR grid & $\{-5,-3,\ldots,19,21\}$~dB \\
Number of echoes $K$ & $\{1, 3\}$ (single echo or multi-echo) \\
Delay $\tau$ & $1$--$33$ samples ($15$~m per sample) \\
Normalized Doppler $f_D$ & echoes $\le8\times10^{-5}$; direct $\le1\times10^{-5}$ \\
Direct fading & Rician, $\kappa = 10$~dB \\
Echo amplitude & geometry, $1/d^4$ power \\
\bottomrule
\end{tabular}

\medskip
\parbox{\columnwidth}{\scriptsize $f_c$: carrier; $f_s$: sampling rate; $N_{\mathrm{seq}}$: burst length in samples, at the $10\,\mu$s coherent time; $K$: number of echoes, one obstacle plus $K-1$ clutter taps; $\tau$: discrete delay in samples; $f_D$: normalized Doppler in cycles/sample; $\kappa$: Rician factor; the echo amplitudes follow the geometry of Sec.~\ref{subsec:scene}.}
\end{table}
 The classifier covers all $33$ delays, while the scene places the labeled obstacle between samples $3$ and $27$. The $33$-sample window is $\approx3.3\,\mu$s of round-trip delay, i.e. $\approx500$~m of one-way range at $f_s{=}10$~MHz, and the echo-Doppler bound of $8\times10^{-5}$ cycles/sample is $\approx800$~Hz, the normalization edge that corresponds to $\approx50$~m/s of relative radial speed at $2.4$~GHz, while the swarm scene samples the typical regime of $10$--$30$~m/s. The direct path uses a tighter bound ($1\times10^{-5}$) and the multi-echo, reduced-Doppler scenario caps the echo bound at $4\times10^{-5}$.

\subsection{Swarm Geometry and Cooperative Peers}
\label{subsec:scene}

The scene replaces the synthetic amplitude prior with a physical one: node altitude $50$~m, obstacle reflectivity $0.30$, reference range $100$~m, and obstacle distances uniformly drawn over $40$--$400$~m, i.e. delays from $3$ to $27$ samples. The obstacle echo follows the $1/d^4$ law and the two clutter taps follow the same geometry at their own ranges. Two consequences shape the whole sensing problem: the echo amplitude spans almost two orders of magnitude over the window, and the distance-to-delay mapping covers the $33$-sample window with $15$~m per sample. Over a $200{,}000$-burst sample of the generator the obstacle is the peak of the matched-filter profile in $45.5\%$ of the bursts, and its mean normalized amplitude is $0.41$; the radial speeds of $10$--$30$~m/s map to echo Doppler shifts of $160$--$480$~Hz.

A drone swarm is not a point target. The receiver of a node also sees its cooperative peers, and because a peer can be much closer than the obstacle, its return is often stronger than the one that carries the range label. In the multi-echo + peers scenario two peers are placed at $10$--$120$~m, with amplitude $\alpha_p=\min\{0.25\,(30/d_p)^2,\,0.6\}$ and Doppler up to half the echo bound. The peer delays are \emph{known} a priori, because the swarm shares a formation plan and a coarse position service, so they are excluded from the range label by construction and the sensing target remains the delay of the dominant obstacle echo. When the obstacle and a peer fall in the same range cell the two returns are indistinguishable within a single burst, so the receiver does not report a range at all: it raises a \emph{fail-safe} whenever the peer-exclusion guard of Sec.~\ref{subsec:gate} is not satisfied.

\subsection{Threat Model and Frequency-Hopping Protocol}
\label{subsec:threat_model}

The nominal aerial multi-echo link of \eqref{eq:multitarget}, with the parameters collected above, is used throughout the paper; the four out-of-distribution propagation models are defined in Sec.~\ref{subsec:ood_channels}. The hopping protocol divides the band into $N_{\mathrm{ch}}=8$ channels and transmits $D$ bursts per channel on a $10\,\mu$s slot, so that the hop rate is $1/(D\cdot10\,\mu\mathrm{s})$ and the dwell sweep $D\in\{1,2,4,8,16\}$ spans $100$ down to $6.25$~khop/s. The jammer is a continuous-wave tone or a partial-band signal occupying $25\%$ of the band, and for the agility study it follows one of four archetypes of the standard taxonomy \cite{poisel2011_jamming}: \emph{barrage}, which covers the whole band and therefore every hop; \emph{fixed-partial}, which covers $25\%$ of the channels; \emph{sweep}, which visits the channels blindly and therefore meets one hop in eight; and \emph{follower}, a reactive follower that observes the hop pattern and engages a channel after a reaction time of $R$ bursts. The quantity that governs the outcome is the jammed fraction $f$, i.e. the share of bursts that meet the jammer: $f{=}1$ for barrage, $f{=}0.25$ for fixed-partial, $f{=}0.125$ for sweep (the measured exposure is $12.2\%$), and, for the follower, a function of both $D$ and $R$.

\section{Proposed Architecture: Ultra-CAN}
\label{sec:architecture}

\begin{figure*}[!t]
\centering
\includegraphics[width=\textwidth]{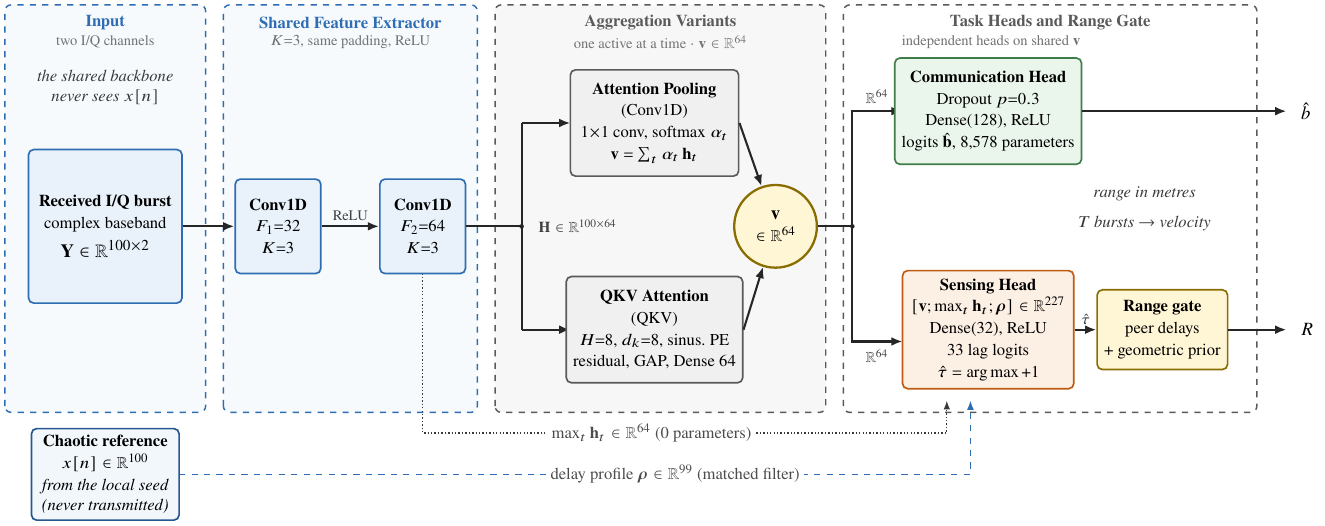}
\caption{Proposed Ultra-CAN dual-mode ISAC receiver: peer-telemetry reception and local-seed monostatic radar share the Conv1D backbone, the two aggregation variants, and the communication, sensing and range-gate heads.}
\label{fig:arch}
\end{figure*}

Fig.~\ref{fig:arch} shows the receiver. The communication stream processes only the two I/Q channels $\mathbf{Y}\in\mathbb{R}^{100\times2}$: because the telemetry link is non-cooperative, the per-symbol chaotic state of the transmitter is not used for demodulation. The sensing stream additionally uses the locally regenerated reference $x[n]$, but that reference never enters the backbone; it is used only to build the delay-profile descriptor of Sec.~\ref{subsec:sensing_head}. This asymmetry is the design point of the paper: one shared feature extractor sees exactly what a blind listener sees, while the node's own seed closes the sensing loop without any over-the-air exchange.

\subsection{Shared Backbone and Aggregation Variants}
\label{subsec:backbone}

The backbone is two one-dimensional convolutions with $F_1{=}32$ and $F_2{=}64$ filters, kernel $K{=}3$, same padding and ReLU. With a $100$-sample burst the feature map is $\mathbf{H}\in\mathbb{R}^{100\times64}$; no pooling is applied before the aggregation, so the temporal resolution of the burst is preserved up to the read-out. Two aggregation variants are compared, both producing the same $64$-dimensional vector $\mathbf{v}$ that the heads consume.

\textbf{Attention pooling (Ultra-CAN).} A $1\times1$ convolution projects $\mathbf{H}$ onto $64$ channels; a dense score $e_t$ is computed for each time position and normalized over the burst, $\alpha_t=\exp(e_t)/\sum_{t'}\exp(e_{t'})$; the representation is the weighted average $\mathbf{v}=\sum_t\alpha_t\mathbf{h}_t$. The softmax makes the read-out selective: the positions that carry the echo structure dominate, and the ones that carry only noise are suppressed.

\textbf{Multi-head QKV attention (Ultra-CAN-QKV).} A scaled dot-product attention with $H{=}8$ heads of dimension $d_k{=}8$ operates on $\mathbf{H}$ with sinusoidal positional encoding and a residual connection; a global average pooling and a dense projection then produce $\mathbf{v}\in\mathbb{R}^{64}$. By capturing all-to-all temporal dependencies across the $100$-sample burst, this multi-head variant provides consistently superior demodulation accuracy across all four evaluated scenarios (Sec.~\ref{subsec:comm}).

Both variants are equally light: $27{,}620$ and $44{,}195$ float32 parameters, i.e. $107.9$ and $172.6$~KiB, with the same heads and the same $64$-dimensional interface. They are two implementations of one design, not two systems. Both run on the same two-layer Conv1D backbone and differ only in the aggregation stage; for brevity, the tables and the figures label the attention-pooled variant \emph{Conv1D}, after the shared backbone, and the multi-head variant \emph{QKV}.

\subsection{Communication Head}
\label{subsec:comm_head}

The vector $\mathbf{v}$ is regularized by dropout ($p{=}0.3$) and fed to a dense layer of $128$ ReLU units followed by the two logits of the blind binary classifier. The head counts $8{,}578$ parameters and is trained with sparse categorical cross-entropy and label smoothing $0.1$. It never receives $x[n]$: the reported communication results are what a blind receiver can decode from the waveform alone. While an idealized coherent oracle given $x[n]$ achieves error-free decoding on the same bursts, the blind heads operate strictly from the received waveform, paying only a modest BER floor ($2.5$--$4.0\times10^{-3}$ across $5$--$21$~dB on the multi-echo link) to eliminate any over-the-air state exchange.

\subsection{Sensing Head and Delay-Profile Descriptor}
\label{subsec:sensing_head}

The sensing head does not regress a continuous delay. It classifies the lag, because the labeled target of the campaign is itself a discrete tap and because a distribution over lags gives the gate of Sec.~\ref{subsec:gate} the confidence it needs.

Three descriptors are concatenated. The first is the shared representation $\mathbf{v}\in\mathbb{R}^{64}$, so the sensing decision rests on the same features that decode the bit. The second is the max-pooled feature map $\max_t\mathbf{h}_t\in\mathbb{R}^{64}$, which preserves the strongest activation of each channel irrespective of its position. The third is a delay profile computed from the received burst and the locally regenerated reference, without Doppler compensation: the direct-path component is estimated by a least-squares projection onto the known reference, $\hat{c}=\sum_n x[n]y[n]/\sum_n x[n]^2\in\mathbb{C}$ --- a single complex coefficient that absorbs the residual carrier phase, which suffices because $f_{D,c}\le10^{-5}$ cycles/sample rotates the direct path by only $2\pi\times10^{-3}$ rad ($0.36^\circ$) over the $100$-sample burst --- and subtracted; the residual is then correlated with $x[n]$ at each lag $\tau\in[1,33]$, stacking the magnitude, real and imaginary parts (99 values). This is the matched-filter front end of a classical delay estimator, but here it is an \emph{input} to the head rather than the decision rule. The concatenated descriptor $[\mathbf{v};\max_t\mathbf{h}_t;\boldsymbol{\rho}]\in\mathbb{R}^{227}$ is processed by a $32$-unit ReLU layer and a linear layer of $33$ logits; the estimate is $\hat{\tau}=\arg\max$ of the logits plus one. The head is trained with sparse categorical cross-entropy against the geometric obstacle lag, so it learns to reject clutter and peer taps that are stronger than the obstacle but do not carry the label.

\subsection{Range Gate and Velocity}
\label{subsec:gate}

A delay estimate is only useful if it is safe to report. Before a range is emitted, the receiver applies the peer-exclusion gate of Fig.~\ref{fig:arch} (the range gate): for every peer whose position is known from the swarm plan, the measured profile at the expected peer delay is compared with the amplitude predicted by the $1/d^2$ link model, and the burst is declared \emph{silent} if any peer return exceeds its prediction by the guard factor $\gamma$. The gate is a single scalar knob: a larger $\gamma$ accepts more bursts and is less cautious. The range is then $R=c\hat{\tau}/(2f_s)$, i.e. $15$~m per sample, and no metric is computed on a silent burst, so the reported error is always the error of the bursts the receiver chose to answer.

Velocity is not an output of a single burst: the $10\,\mu$s coherent time cannot resolve the $160$--$480$~Hz echo Doppler. In radar mode it is recovered at system level as the slope of a track of successive range estimates, which costs nothing because the receiver is already producing one range per burst.

\subsection{Multi-Task Loss}
\label{subsec:loss}

The two heads are trained end-to-end with
\begin{equation}
\mathcal{L} = \mathcal{L}_{\mathrm{comm}} + \lambda\,\mathcal{L}_{\mathrm{lag}},
\label{eq:mtl_loss}
\end{equation}
where $\mathcal{L}_{\mathrm{comm}}$ is the label-smoothed cross-entropy of the bit, $\mathcal{L}_{\mathrm{lag}}$ is the sparse categorical cross-entropy over the $33$ lag logits, and $\lambda{=}2$ is a fixed scalar weight. The value was selected by a validation-set sensitivity check: smaller weights underfit the sensing head, while larger weights let the sensing term dominate the total loss. With $\lambda{=}2$ the sensing task stays active without competing with the decoder, and the communication results of Sec.~\ref{subsec:comm} are obtained with the shared, multi-task model rather than with a communication-only variant; this is what licenses the sharing claim of the paper. No adaptive weighting \cite{kendall2018_mtl_loss, chen2018_gradnorm} is needed at this scale, and a second Doppler output is not trained because the single-burst Doppler target is not resolvable.

\section{Performance Evaluation}
\label{sec:evaluation}

\subsection{Experimental Setup}
\label{subsec:setup}

\begin{figure*}[!t]
\centering
\includegraphics[width=\textwidth]{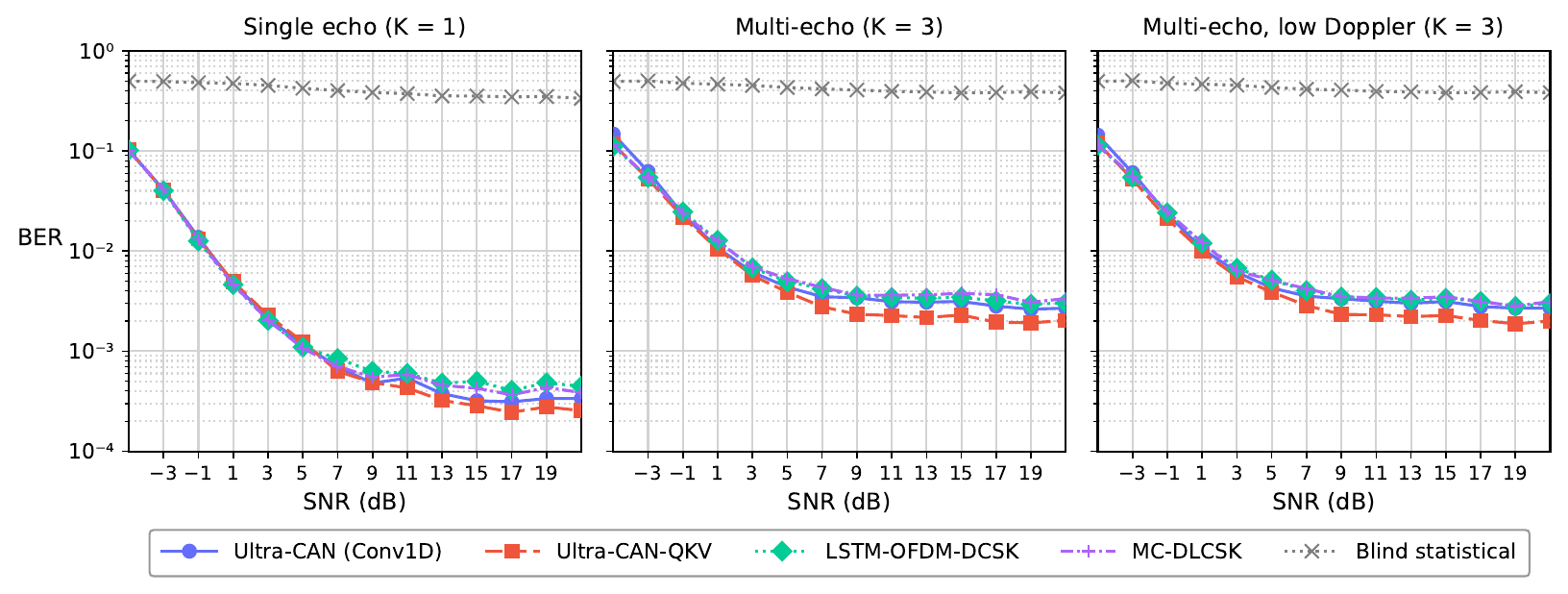}
\caption{BER against SNR over the full grid: (a) single echo, (b) multi-echo, (c) multi-echo, reduced Doppler. The blind statistical receiver is included; the proposed variants are the blue and red curves.}
\label{fig:ber}
\end{figure*}

To evaluate the receivers across the SNR operating region (Fig.~\ref{fig:ber}), training and validation datasets are generated offline using the channel model of \eqref{eq:multitarget} and the link parameters of Sec.~\ref{subsec:channel_model}. For each architecture and scenario a single model is trained on a set whose SNR is uniform over the grid, so every per-SNR curve is produced by one model with no per-SNR retraining. Training uses $105{,}000$ symbols and $5{,}000$ validation symbols; testing follows the ``at least $100$ bit errors'' criterion at each SNR point, with a cap of $2\times10^{6}$ symbols and online generation. All receivers use Adam ($10^{-3}$), batch $128$, at most $30$ epochs with early stopping (patience $10$), a cosine schedule and label smoothing $0.1$. The LSTM-OFDM-DCSK and MC-DLCSK baselines are re-implemented from \cite{zhang2020, mobini2025} within the same TensorFlow/Keras stack, dataset and procedure, and their communication stage also receives only the two I/Q channels, so all neural receivers are blind and their parameter counts refer to these re-implementations. A blind statistical receiver is evaluated on the same test points as a classical non-coherent reference: it projects each burst onto the principal axis of its $2\times2$ I/Q covariance and classifies five amplitude-shape statistics of the normalized projection (mean magnitude, two tail fractions, kurtosis and skewness) with a linear discriminant fitted on the training split, so it needs neither the chaotic reference nor any training beyond that fit.

We report two operating-region metrics: the pooled BER over $\mathrm{SNR}\ge5$~dB (the mean of the per-point BERs, as in the released tables) and the best per-point BER inside that region. The gross bit rate follows from the burst: one bit per $100$-sample block at $f_s{=}10$~MHz is $100$~kbit/s, halved in the half-duplex telemetry mode. With the $100$-error rule the per-point $1\sigma$ relative uncertainty is $\approx10\%$, so differences below that level are not resolved.

\textbf{Out-of-distribution channels.}\label{subsec:ood_channels} The cross-channel study uses four propagation models outside the training distribution: the 3GPP TDL-D profile with $25\%$ and with $5\%$ diffuse clutter, the TDL-A profile, and a two-ray geometry with Jakes fading. Both TDL profiles model an aerial link with a $300$~ns RMS delay spread \cite{khawaja2019_uav_channel}, so that $11$ to $12$ taps fall inside the $33$-sample window, and they differ in whether a specular (LOS) component is present. Unlike the nominal link, where the channel is redrawn symbol by symbol, the profiles assume a per-hop coherence block of $53$ slots. The two-ray variant places a single specular reflector at an excess delay of $167$~ns ($1.67$ samples) carrying $21.8\%$ of the received power, with a Doppler of $160$~Hz.

\subsection{Telemetry Demodulation Performance}
\label{subsec:comm}

\textbf{Classical baselines.}\label{subsec:classical} The case for a learned receiver is only meaningful against the classical receivers a chaos-based radio would otherwise deploy. We compare three of them --- the differential DCSK correlator, a matched filter against the two nominal map signatures, and an energy detector --- each with at most one decision threshold fitted on the training split, evaluated on a deliberately idealized link (echoes only, no additive noise, no echo fading) so that the comparison is as favorable to the classical chain as possible. The energy detector sits at the coin-flip level in every condition, which confirms that the modulation carries its information in the waveform shape and not in its power. The DCSK correlator stays at BER $0.27$--$0.33$ even with a single echo and does not improve as echoes are added. Only the matched filter retains a usable margin ($2.4\times10^{-2}$ with one echo), and it degrades monotonically with the echo Doppler, reaching $0.41$ at the edge of the grid, approaching the coin-flip level, because the Doppler rotation destroys the fixed template. The classical chain therefore fails on exactly the aerial link this paper targets (Table~\ref{tab:classical}), and it fails without the noise the learned receiver has to fight.

\begin{table}[!b]
\centering
\caption{Classical baselines on the noise-free link.}
\label{tab:classical}
\footnotesize
\begin{tabular}{lccc}
\toprule
Condition & Correlator & Matched filter & Energy det. \\
\midrule
$K=1$ & $0.269$ & $0.024$ & $0.50$ \\
$K=5$ & $0.293$ & $0.056$ & $0.50$ \\
$K=10$ & $0.329$ & $0.098$ & $0.50$ \\
\midrule
$f_D{=}0$ & $0.283$ & $0.039$ & $0.50$ \\
$f_D{=}1.1\times10^{-3}$ & $0.304$ & $0.072$ & $0.50$ \\
$f_D{=}3.0\times10^{-3}$ & $0.374$ & $0.406$ & $0.50$ \\
\bottomrule
\end{tabular}

\medskip
\parbox{\columnwidth}{\scriptsize $K$: number of echoes; $f_D$: normalized Doppler in cycles/sample; correlator: differential DCSK; matched filter: correlation against the two nominal map signatures; energy det.: energy detector. The link carries echoes only, with no additive noise and no echo fading.}
\end{table}

\textbf{Learned receivers.} Fig.~\ref{fig:ber} reports the BER of the four learned receivers and of the blind statistical receiver over the whole grid, for the single-echo, multi-echo and multi-echo, reduced-Doppler scenarios. Table~\ref{tab:metrics} collects the pooled values, including the multi-echo + peers scenario.

\begin{table}[!t]
\centering
\caption{Pooled BER; best in bold.}
\label{tab:metrics}
\footnotesize
\setlength{\tabcolsep}{2pt}
\begin{tabular}{@{}lcccc@{}}
\toprule
Scenario & Conv1D & QKV & LSTM & MC-DLCSK \\
\midrule
Single echo & $5.08{\times}10^{-4}$ & $\mathbf{4.64{\times}10^{-4}}$ & $6.10{\times}10^{-4}$ & $5.53{\times}10^{-4}$ \\
Multi-echo & $3.20{\times}10^{-3}$ & $\mathbf{2.41{\times}10^{-3}}$ & $3.57{\times}10^{-3}$ & $3.83{\times}10^{-3}$ \\
Reduced Doppler & $3.18{\times}10^{-3}$ & $\mathbf{2.42{\times}10^{-3}}$ & $3.51{\times}10^{-3}$ & $3.58{\times}10^{-3}$ \\
Multi-echo $+$ peers & $2.37{\times}10^{-3}$ & $\mathbf{1.76{\times}10^{-3}}$ & $4.21{\times}10^{-2}$ & $4.58{\times}10^{-3}$ \\
\bottomrule
\end{tabular}

\medskip
\parbox{\columnwidth}{\scriptsize single echo: $K{=}1$, all other rows $K{=}3$; reduced Doppler: multi-echo with the echo-Doppler bound halved. pooled BER: mean of the per-point BERs over the nine SNR points $\ge5$~dB; LSTM: LSTM-OFDM-DCSK; MC-DLCSK: multilevel code-index DCSK; both re-implemented in the same stack; best in bold.}
\end{table}

The QKV variant is the best demodulator in all four scenarios and the Conv1D variant is second, so the proposed pair occupies the first two places of the learned ranking in the point estimates. On the single-echo link the QKV variant reaches a pooled BER of $4.64\times10^{-4}$, $16\%$ below the best baseline, and a best per-point BER of $2.46\times10^{-4}$ at $17$~dB. On the multi-echo link it is $32\%$ below the best baseline ($2.41\times10^{-3}$ against $3.57\times10^{-3}$), and on the multi-echo + peers link the margin grows to $62\%$ ($1.76\times10^{-3}$ against $4.58\times10^{-3}$): cooperative peers are extra multipath, and the learned front end absorbs them better than the recurrent baselines. The multi-echo, reduced-Doppler scenario confirms that the ranking does not depend on the Doppler bound ($2.42\times10^{-3}$ against $3.51\times10^{-3}$). The LSTM-OFDM-DCSK entry is the outlier in Table~\ref{tab:metrics}: its pooled BER is $4.21\times10^{-2}$ in the multi-echo + peers scenario, against $3.6\times10^{-3}$ in the peer-free multi-echo case. The value is a measured failure of the recurrent baseline under peer interference ($3.93\times10^{-2}$ at $21$~dB on the same evaluation grid), not a transcription error, and we retain it as a result. The single-echo ordering is also the weakest statistically: the QKV/Conv1D gap there is $4.4\times10^{-5}$ in pooled BER with a paired $95\%$ interval of $[0.8,7.3]\times10^{-5}$ across the nine operating points, whereas the multi-echo and peer-laden gaps separate. The ranking in the abstract refers to these point estimates.

The blind statistical receiver, which requires no training, stays two to three orders of magnitude above every learned receiver ($0.369$ single-echo and $0.396$ multi-echo pooled), so the learned front end is what makes the operating region reachable in the first place. The severe $1/d^4$ geometric attenuation bounds all receivers above a $10^{-4}$ floor, with Ultra-CAN-QKV achieving the lowest per-point BER at $2.46\times10^{-4}$. When the channel is relaxed to a log-uniform echo-amplitude prior, the Conv1D and QKV variants reach per-point BERs of $1.10\times10^{-4}$ and $1.68\times10^{-4}$, confirming that the error floor is governed by physical echo dynamics rather than model capacity. The same ablation raises the model-free peak fraction from $0.455$ to $0.855$, i.e. the obstacle becomes the strongest tap in most bursts, the quantity that governs the sensing result of Sec.~\ref{subsec:sensing}.

\subsection{Radar-Mode Ranging, Peer-Aware Gating, and Velocity Tracking}
\label{subsec:sensing}

\textbf{Obstacle identification.} Beyond bit demodulation, Ultra-CAN extracts the obstacle range from the radar burst through the delay-profile descriptor of Sec.~\ref{subsec:sensing_head}, outperforming correlation and 5G-NR OFDM-ISAC baselines at $21$~dB. The metric is the exact identification of the labeled obstacle lag, with a tolerance of $0.5$ samples: the estimate is exact when it falls on the obstacle tap, and it is wrong when it falls on one of the clutter or peer taps. The quantity is deliberately harder than a range error, because it asks \emph{which} return the receiver reports.

The measured rates at $21$~dB on the multi-echo + peers scenario, the operating point of the swarm, are as follows. A chaotic matched filter, the model-free baseline of the same waveform, identifies the obstacle in $42.4\%$ of the bursts; a 5G-NR OFDM-ISAC processor with an extended cyclic prefix and the full pilot grid reaches $44.2\%$ ($44.3\%$ with the $15$~kHz numerology) --- both at the same $10$~MHz bandwidth and energy per burst ($168$ subcarriers at $60$~kHz, $660$ at $15$~kHz), with perfect synchronization and a single OFDM symbol, without slot integration or Doppler --- and the empirical peak-picker ceiling --- the share of bursts in which the obstacle is the strongest non-peer tap --- is $48.0\%$. This ceiling is not an upper bound on a classifier: it is the rate a correlation-only receiver can reach. The Ultra-CAN head reaches $66.8\%$, $18.8$ points above that ceiling and $58\%$ above the matched filter. The two variants agree to two tenths of a point ($66.6\%$ for the QKV variant), so the gain is a property of the supervision and of the shared representation, not of the attention mechanism. Over the same $200{,}000$ bursts the absolute lag error has a median of $0$, a $75$th percentile of $7$ samples and a $90$th percentile of $16$ samples ($240$~m), and a mean of $3.97$ samples for Conv1D ($3.95$ for QKV).

The same breakdown shows \emph{where} the remaining reports land, which is the part a range sensor used for formation keeping cares about. Because the label is the geometric obstacle, every report is either the obstacle, a peer, or a clutter tap. The two correlation-only receivers put one burst in six on a peer ($17.4\%$ and $17.9\%$) and about $40\%$ on clutter, so they are wrong on more than half of the bursts they answer. The head reduces the peer reports to $4.4\%$ and the clutter reports to $28.7\%$, i.e. it reports the tap that the label defines in two bursts out of three. The error distribution follows from that composition and is bimodal by construction: it is exactly zero when the head reports the obstacle, and it is displaced by the distance to another tap when it does not, which is why the mean sits two orders of magnitude above the median (Table~\ref{tab:sensing}).

\begin{table}[!t]
\centering
\caption{Sensing outcome at $21$~dB on $200{,}000$ peer-laden bursts.}
\label{tab:sensing}
\footnotesize
\setlength{\tabcolsep}{2.5pt}
\begin{tabular}{@{}lcccc@{}}
\toprule
 & \multicolumn{3}{c}{reported tap [\% of answered bursts]} & \\
\cmidrule(lr){2-4}
Method & obst. & peer & clutter & ans. \\
\midrule
Chaotic MF & $42.4$ & $17.4$ & $40.2$ & $100$ \\
OFDM-ISAC, $60$~kHz & $44.2$ & $17.9$ & $37.9$ & $100$ \\
OFDM-ISAC, $15$~kHz & $44.3$ & $17.8$ & $37.9$ & $100$ \\
Peak-picker ceiling & $48.0$ & --- & --- & $100$ \\
Ultra-CAN (Conv1D) & $66.8$ & $4.4$ & $28.7$ & $100$ \\
Ultra-CAN-QKV & $66.6$ & $4.4$ & $29.0$ & $100$ \\
\midrule
\multicolumn{5}{@{}l}{\emph{with the peer-exclusion gate, $\gamma{=}1.5$}} \\
Ultra-CAN (Conv1D) & $74.9$ & $0.02$ & $25.1$ & $23.5$ \\
Ultra-CAN-QKV & $75.1$ & $0.03$ & $24.9$ & $23.5$ \\
\bottomrule
\end{tabular}

\medskip
\parbox{\columnwidth}{\scriptsize obst., peer, clutter: share of the answered bursts whose reported tap is the labeled obstacle, a peer echo, or any other tap, so obst.\ is the precision of the answered list; ans.: share of the $200{,}000$ bursts that receive a range ($100$ for the baselines and for the ungated head, which answer every burst). The ceiling is defined on the non-peer taps, so it has no peer or clutter entry.}
\end{table}

The gain is not specific to this scenario. On the peer-free multi-echo link, over the same $200{,}000$ bursts, the head identifies the obstacle in $69.4\%$ (Conv1D) and $69.8\%$ (QKV) of the cases, against $45.6\%$ for the correlation peak measured on the same bursts; on the single-echo link both reach $100\%$ above $5$~dB. What the head adds over the profile is a read-out that has learned the \emph{geometry} of the scene: the obstacle is not always the strongest return, but it has a signature in the complex profile --- a position, a width and a phase relation with the direct path --- that the $33$-way classifier can exploit. The model-free profile remains the input, so the sensing function still rests on a classical descriptor; the difference is the learned decision on top of it. This is the multi-task result the paper claims: the sensing head does not need a dedicated radar chain, and it beats the peak picker that a correlation-only receiver would use. The head rows are ungated; applying the peer-exclusion gate evaluated below suppresses the peer reports to $0.02\%$ of the answered bursts while three answered ranges in four remain on the exact obstacle tap.

The correlation coefficient between the estimated and the true delay is $0.486$ (Conv1D) and $0.493$ (QKV) on the nominal link, and $0.443$ for both on the multi-echo + peers link. It is lower in the multi-echo + peers scenario by construction: the label is the geometric obstacle, and in half of the bursts the obstacle is not the strongest return, so a linear correlation with the delay understates a classifier that reports the correct tap in two bursts out of three.

\begin{figure*}[!t]
\centering
\includegraphics[width=\textwidth]{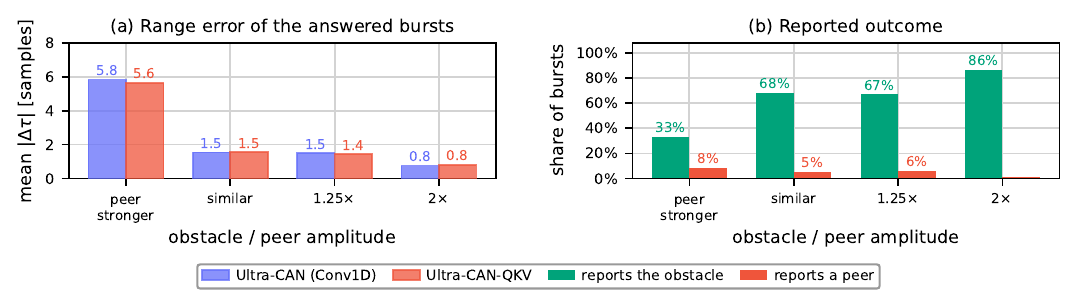}
\caption{Ranging on $200{,}000$ peer-laden bursts at $21$~dB: (a) mean range error of the answered bursts and (b) reported outcome, against the obstacle-to-peer echo ratio.}
\label{fig:peers}
\end{figure*}

\textbf{Peer-exclusion gate.}\label{subsec:peers_eval} The peer-aware mode turns the lag estimate into a safe range by excluding cooperative peers before the range is emitted. Without the gate, the head reports a peer as the obstacle in $4.4\%$ of the bursts, which is the failure a range sensor used for formation keeping must never produce. On the bursts it answers, the estimate falls on the exact obstacle lag in $74.9\%$ of the cases ($75.1\%$ for QKV) and the peer reports drop from $4.4\%$ of all bursts to $0.02\%$ of the answered ones. Fig.~\ref{fig:peers}a--b summarizes the range error and the reported outcome over $200{,}000$ peer-laden bursts at $21$~dB; at $\gamma{=}1.5$ the gate answers $23.5\%$ of the bursts.

In the $200{,}000$-burst breakdown of Fig.~\ref{fig:peers}, the mean error is $5.8$ samples ($5.6$ for QKV) when the peer is stronger than the obstacle, falls to $1.5$ when the two echoes are comparable and to $0.8$ when the obstacle is at least twice as strong. The composition of the reports follows the same ordering: the obstacle is reported in $33\%$ of the bursts when the peer is stronger, in $68\%$ when the two are comparable, and in $67\%$ and $86\%$ when the obstacle is $1.25$ and $2$ times stronger; the ungated peer-report rate falls from $8.5\%$ to $1.5\%$ over the same bins.


Abstention, not peer reporting, is the knob that sets the operating point. In the $200{,}000$-burst gate sweep, the answered fraction rises with the guard factor $\gamma$ from $13.1\%$ at $\gamma{=}1$ to $32.0\%$ at $\gamma{=}2$, while among the answered bursts the share in which a peer is reported as the obstacle is at most $0.05\%$ at every guard, against the $4.4\%$ of the ungated head. The precision of the answered list moves much less than its length: $76.8\%$ of the answered ranges land on the obstacle at $\gamma{=}1$, $74.9\%$ at $\gamma{=}1.5$ and $73.5\%$ at $\gamma{=}2$, because what the gate admits is mostly the static clutter that no geometric prior can predict. The two architectures take identical gate decisions, since the gate is a deterministic test on the profile.

When the obstacle and a peer share the same range cell the gate trips in $100\%$ of the cases: a single burst cannot separate two returns in the same delay cell, and the receiver says so instead of guessing.

At the default guard $\gamma{=}1.5$ the receiver answers $23.5\%$ of the bursts, reports the obstacle in $74.9\%$ of the answered ones and a peer in $0.02\%$ of them. The operating point is therefore a deployment parameter: a formation controller that needs a range every few bursts can relax $\gamma$, whereas a collision-avoidance layer that cannot tolerate a wrong range keeps it tight. The gate needs only the nominal peer positions, not their echoes: the formation plan that the swarm already shares supplies the expected delays, and the guard factor absorbs the residual uncertainty without an extra message. The cost is a shorter answer list, and the failure mode it removes is the one that matters in a cooperative swarm: a peer reported as an obstacle. Static clutter survives the gate, because the formation plan predicts the peers and not the terrain, which is why the gate is a fail-safe against peer-induced ambiguity rather than a certificate on the obstacle map.

\textbf{Velocity tracking.}\label{subsec:velocity} A single $10\,\mu$s burst cannot resolve the $160$--$480$~Hz echo Doppler, so velocity is not an output of the receiver but the slope of a track of successive range estimates. 

\begin{figure}[t]
\centering
\includegraphics[width=\columnwidth]{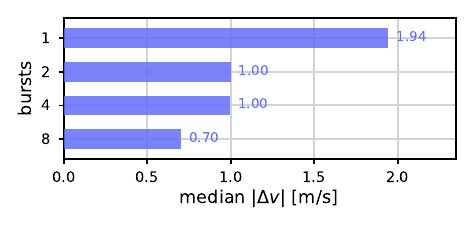}
\caption{Median absolute velocity error of the model-free profile tracker against the number of bursts averaged into the track, over thirty tracks of $50$ bursts.}
\label{fig:velocity}
\end{figure}

The tracker is estimator-agnostic; the reported error uses the model-free profile peak, not the sensing head, because the head is trained for exact tap identification, a discrete decision, whereas velocity needs sub-sample delay differences. The two outputs are complementary: the head answers which return, the profile interpolation answers how much the range moved. A plain moving average over the delay sequence gives a median absolute velocity error over thirty tracks of $50$ bursts that falls from $1.94$~m/s for a single-burst difference to $0.70$~m/s over eight bursts, against true radial speeds of $10$--$30$~m/s. The experiment runs at $15$~dB, and the error falls as the window grows because the range noise of successive bursts averages out while the closing speed stays constant across the window; the estimate is the radial closing speed, not the full velocity vector, which is what formation keeping and collision avoidance need. Eight bursts, i.e. a $0.7$~s window at the $100$~ms cadence used in this experiment ($1.25$ non-overlapping updates per second), already place the estimate within a meter per second, the closing-speed resolution a swarm needs, and it comes from bursts the receiver was already decoding. A shorter window is possible when the eight bursts are acquired back-to-back; Sec.~\ref{subsec:mcu} reports the rate that this implies. Fig.~\ref{fig:velocity} illustrates the tracking error.

\begin{table}[!t]
\centering
\caption{Cross-channel; best in bold.}
\label{tab:generalization}
\footnotesize
\renewcommand{\arraystretch}{0.95}
\setlength{\tabcolsep}{3pt}
\begin{tabular}{@{}lccccc@{}}
\toprule
Channel & $P_{\mathrm{peak}}$ & \multicolumn{2}{c}{pooled BER} & \multicolumn{2}{c}{$\rho_\tau$} \\
\cmidrule(lr){3-4}\cmidrule(lr){5-6}
 & & Conv1D & QKV & Conv1D & QKV \\
\midrule
nominal & $0.455$ & $3.20{\times}10^{-3}$ & $\mathbf{2.41{\times}10^{-3}}$ & $0.486$ & $\mathbf{0.493}$ \\
TDL-D (LOS) & $0.026$ & $3.77{\times}10^{-1}$ & $\mathbf{3.60{\times}10^{-1}}$ & $\mathbf{0.008}$ & $0.007$ \\
TDL-D, $5\%$ clutter & $0.026$ & $3.35{\times}10^{-1}$ & $\mathbf{3.12{\times}10^{-1}}$ & $0.009$ & $\mathbf{0.011}$ \\
TDL-A (NLOS) & $0.026$ & $3.84{\times}10^{-1}$ & $\mathbf{3.70{\times}10^{-1}}$ & $\mathbf{0.009}$ & $\mathbf{0.009}$ \\
two-ray $+$ Jakes & $0.027$ & $3.38{\times}10^{-1}$ & $\mathbf{3.17{\times}10^{-1}}$ & $0.006$ & $\mathbf{0.012}$ \\
TDL-D, retrained & $0.026$ & $\mathbf{5.68{\times}10^{-3}}$ & $5.90{\times}10^{-3}$ & $0.018$ & $\mathbf{0.019}$ \\
\bottomrule
\end{tabular}

\medskip
\parbox{\columnwidth}{\scriptsize $P_{\mathrm{peak}}$: model-free probability that the labeled obstacle is the strongest non-peer tap of the profile; $\rho_\tau$: correlation between the estimated and the true lag; pooled BER over $\mathrm{SNR}\ge5$~dB; the nominal row is the peer-free multi-echo link of Sec.~\ref{subsec:comm}; the last row is the receiver retrained on the same link.}
\end{table}

\subsection{Cross-Channel Generalization and Edge Adaptation}
\label{subsec:generalization}

An edge receiver is practically useful only if its behavior under unseen propagation conditions is predictable and recoverable (Table~\ref{tab:generalization}). The table reports pooled BER over $\mathrm{SNR}\ge5$~dB; the BER floor quoted below is the minimum over the grid, and the nominal row repeats the peer-free multi-echo pooled BER of Sec.~\ref{subsec:comm}, while the last row is the receiver retrained in domain.

 We first evaluate the two proposed variants with frozen weights across the four out-of-distribution models of Sec.~\ref{subsec:ood_channels}, together with the model-free probability $P_{\mathrm{peak}}$ that the labeled obstacle is the strongest non-peer tap of the profile. Two facts stand out. On the nominal link the label is the profile peak in $45.5\%$ of the bursts; on the TDL profiles and on the two-ray geometry it is the peak in $2.6$--$2.7\%$ only, because the diffuse clutter carries most of the received power.

 The frozen receivers follow that drop: their pooled BER moves to $0.31$--$0.38$ and their delay correlation to $0.006$--$0.012$.

The QKV variant remains the better demodulator everywhere ($3$--$7\%$ lower pooled BER), but the gap is no longer the $25\%$ of the nominal link: off the training distribution the two backbones converge, which is the expected signature of a domain shift rather than a property of the architecture.

\begin{figure*}[!t]
\centering
\includegraphics[width=\textwidth]{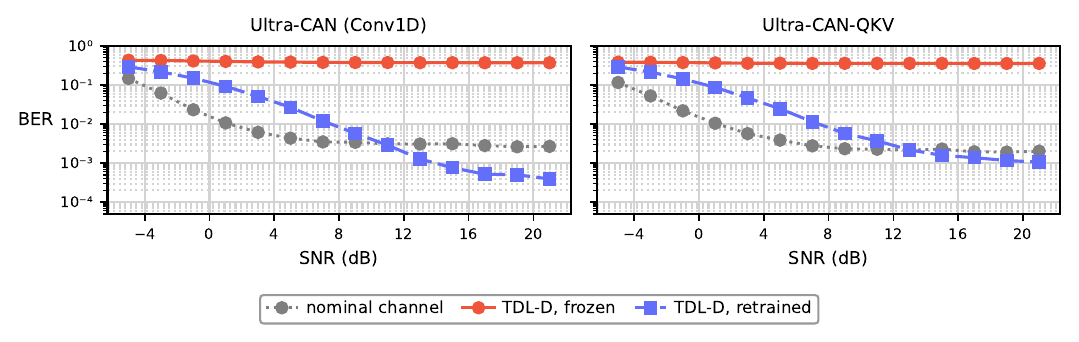}
\caption{Retraining control on the TDL-D link: nominal channel (grey dotted), frozen TDL-D (red solid) and TDL-D retrained on $105{,}000$ symbols (blue dashed), for the two backbones.}
\label{fig:retraining}
\end{figure*}
The mechanism is visible in the table: the failure is not in the backbone, it is in the mismatch between the channel the receiver learned and the channel it sees. The remedy is in-domain retraining. Refitting each backbone on $105{,}000$ symbols of the TDL-D link lowers the BER floor from $0.372$ to $3.9\times10^{-4}$ for the Conv1D variant ($946\times$) and from $0.358$ to $1.1\times10^{-3}$ for the QKV variant ($337\times$), below the nominal in-domain floors of $2.7\times10^{-3}$ and $2.0\times10^{-3}$.

The recovered curves cross the nominal ones near $13$~dB: at $21$~dB the retrained receivers sit $6.8\times$ and $1.9\times$ below the nominal points, while at $5$~dB they pay $0.027$ and $0.025$ against $4.4\times10^{-3}$ and $3.9\times10^{-3}$, and the pooled BER over $\mathrm{SNR}\ge5$~dB returns $5.7\times10^{-3}$ and $5.9\times10^{-3}$, within a factor $1.9$ and $2.5$ of the nominal values. The parameter budget is unchanged at $27.6$k/$44.2$k, so the domain shift is a property of the weights and not of the model class, and the receiver is light enough to be re-fitted at the edge from a local capture. Fig.~\ref{fig:retraining} compares the three curves of each backbone.

The delay head does not follow: its best correlation over the grid is $0.018$ and $0.019$, two orders of magnitude below the nominal $0.486$ and $0.493$ and consistent with the $2.6\%$ peak observability of the link, confirming that while in-domain retraining fully recovers telemetry demodulation under heavy diffuse multipath, single-burst delay estimation remains physically bounded by the intrinsic tap observability of the channel.

\subsection{Adversarial Resilience and SWaP-C Footprint}
\label{subsec:resilience}

\textbf{Jamming robustness and jamming-aware training.}\label{subsec:jamming} Both proposed receivers are trained without jamming and evaluated on the peer-free multi-echo link at $5$~dB under continuous-wave (CW), barrage and partial-band attacks over a JSR grid from $-10$ to $+10$~dB with five Monte-Carlo realizations per point. At the top of the grid the CW and partial-band attacks drive every receiver toward the chance level ($0.44$--$0.50$), the QKV variant being the most robust under CW ($0.443$, against $0.498$ for the Conv1D and $0.502$ for both recurrent baselines), while barrage degrades more gently ($0.405$--$0.420$). Robustness to localized jamming is therefore not where the architectures separate; what separates them is \emph{where} inside the receiver the damage is done.

The cosine similarity between the clean and the jammed activations at JSR $+10$~dB measures where the damage lands. On the QKV backbone the convolutional features retain $0.78$--$0.93$ of their clean similarity across the three attacks, the delay profile $0.57$--$0.66$ and the energy-normalized sensing descriptor $0.27$--$0.37$. The endpoint is what matters: with a CW jammer $2$~dB stronger than the signal the QKV BER is already $0.401$, while its delay correlation retains $81\%$ of its clean value. A jammed drone therefore loses its telemetry before it loses its reflector, which is the strongest support for the single-receiver claim.

The repair belongs in training rather than in added hardware: retraining the QKV receiver on a set in which half of the symbols are randomly jammed, with the jammer type drawn uniformly among CW, barrage and partial-band and the JSR over the same grid, restores the link where the attack is localized. Table~\ref{tab:jamaware} reports the comparison. Under CW at $-2$~dB the BER falls from $0.356$ to $0.021$ (a factor $17$), under partial-band at $+2$~dB from $0.445$ to $0.081$, and the decision margin (mean logit separation) returns to $2.53$ and $2.01$; the cost on the clean channel is a BER of $0.0048\to0.0077$. Barrage is the control that does not move ($0.388\to0.399$ at $+10$~dB), as expected for an attack that covers the whole band: it cannot be repaired by making the receiver smarter, only by moving the burst away from it, which is the role of frequency agility, evaluated next.

\begin{table}[!htb]
\centering
\caption{Jamming-aware training of the QKV receiver.}
\label{tab:jamaware}
\footnotesize
\setlength{\tabcolsep}{3.5pt}
\begin{tabular}{@{}lcccc@{}}
\toprule
 & \multicolumn{2}{c}{BER} & \multicolumn{2}{c}{margin} \\
\cmidrule(lr){2-3}\cmidrule(lr){4-5}
Condition & clean & aware & clean & aware \\
\midrule
clean & $\mathbf{0.0048}$ & $0.0077$ & $\mathbf{2.87}$ & $2.72$ \\
CW, $-2$~dB & $0.356$ & $\mathbf{0.021}$ & $0.91$ & $\mathbf{2.53}$ \\
CW, $+6$~dB & $0.405$ & $\mathbf{0.110}$ & $0.34$ & $\mathbf{1.63}$ \\
fixed-partial, $-2$~dB & $0.245$ & $\mathbf{0.041}$ & $1.49$ & $\mathbf{2.36}$ \\
fixed-partial, $+2$~dB & $0.445$ & $\mathbf{0.081}$ & $0.41$ & $\mathbf{2.01}$ \\
barrage, $+2$~dB & $\mathbf{0.084}$ & $0.109$ & $\mathbf{2.11}$ & $1.69$ \\
barrage, $+10$~dB & $\mathbf{0.388}$ & $0.399$ & $\mathbf{0.63}$ & $0.31$ \\
\bottomrule
\end{tabular}

\medskip
\parbox{\columnwidth}{\scriptsize clean: trained without jamming; aware: trained with jamming (Sec.~\ref{subsec:jamming}); margin: mean logit separation; CW: continuous-wave tone; fixed-partial: partial-band; JSR as indicated.}
\end{table}

\textbf{Frequency agility.}\label{subsec:agility} Frequency hopping is the natural complement of a wideband, noise-like waveform \cite{boroujeni2026_fh_isac, cheng2025_antijam_fh}, and the single receiver is the natural place to exploit it because it already sees how often it was jammed. The agility campaign runs on the TDL-D profile with per-hop coherence ($53$ slots), i.e. the channel in which a hop changes the channel realization. Against an aligned jammer that camps on a fixed carrier, every burst is exposed ($f{=}1$); with hopping at $100$~khop/s, exposure drops to $25\%$ against a fixed-partial jammer, $12.2\%$ against a blind sweep, and $0\%$ against a reactive follower, whose reaction time exceeds the dwell. The barrage control keeps $f{=}1$: it cannot be out-hopped. Compared with an unaligned static carrier, hopping leaves the blind sweep unchanged ($12.5\%$ against $12.2\%$) while reducing fixed-partial exposure from $40\%$ to $25\%$: its primary value is therefore the geometric protection against an aligned jammer, in which an omniscient jammer meets every burst, while a reactive follower that never engages simply leaves the link clean. Table~\ref{tab:agility} verifies the resulting mixing law at JSR $+10$~dB.

The protection obeys a mixing law: a burst either meets the jammer, in which case it behaves as in the aligned case, or it is clean, so that
\begin{equation}
\mathrm{BER}_{\mathrm{hop}} = f\,\mathrm{BER}_{\mathrm{jam}} + (1-f)\,\mathrm{BER}_{\mathrm{clean}},
\label{eq:mixing}
\end{equation}
with $f$ the jammed fraction and $G\simeq-10\log_{10}f$ the geometric gain over the aligned case. The table verifies \eqref{eq:mixing} against the measurements at JSR $+10$~dB. Over the $192$ conditions of the sweep (four receivers, four archetypes, two in-channel types, six JSR points) the measured BER follows the prediction with a median relative error of $0.11\%$ and a worst case of $1.3\%$. The consequence is the central one: the gain is a property of the jammed fraction, not of the receiver --- it is the same for the four architectures and it would apply to a classical correlator as well. Hopping therefore protects the shared receiver \emph{geometrically}, by moving the burst away from the jammer rather than by making the receiver smarter, and the design rule that follows is that the dwell must be shorter than the jammer's reaction time. The agility results are protocol-level and are obtained on recorded bursts. With $0.148$~ms per burst on the reference CPU the Conv1D receiver processes at most $\approx6.8$k bursts/s, i.e. one burst every $148\,\mu$s, which is a raw throughput: the dwell-limited rate that it supports follows from $D_{\min}=15$ and is $6.67$~kHz; because a dwell holds the channel, a receiver that cannot follow every burst samples one burst per dwell and still sees the same channel, so with $D_{\min}=\lceil t_r/10\,\mu\mathrm{s}\rceil=15$ the $6.25$~khop/s setting ($D{=}16$) is decoded at one burst per hop, i.e. $6.25$~kbit/s of payload, \eqref{eq:rate}. Faster hopping, or the full $100$~kbit/s air rate, requires $t_r\le10\,\mu$s, i.e. a dedicated front end, and is outside the neural receiver's processing loop. The mixing law itself depends on the jammed fraction, not on the receiver latency.

\textbf{Low-probability-of-intercept profile.}\label{subsec:lpi} The secure side of the single-receiver claim is a property of the waveform: it has a spectral flatness of $0.944$ and a peak-to-floor distance of $3.71$~dB, against $0.045$ and $28.1$~dB for the narrowband baselines (BPSK and OOK), so there is neither a carrier line nor a sidelobe to lock onto; its peak-to-average power ratio is $6.91$~dB, far from the $10$--$12$~dB of an OFDM waveform and well inside the linear range of a small amplifier. Against a channelized radiometer at $P_{fa}{=}10^{-2}$ --- the adversary that recovers the conventional waveforms --- the chaotic waveform is detected with $P_d{=}0.017$ at $16$~dB, against $0.764$ for BPSK and $0.842$ for OOK, and it plateaus at $0.297$ even at $24$~dB. Against a plain radiometer, all equal-energy waveforms are indistinguishable. The limit is stated in the clear: conventional DSSS is slightly flatter still ($0.999$), so the advantage holds against narrowband standard modulations and not against a spread-spectrum adversary; and it assumes that the map parameters are not disclosed, which is a key-management assumption rather than an unconditional guarantee.
\begin{table}[!t]
\centering
\caption{Mixing law at JSR $+10$~dB.}
\label{tab:agility}
\footnotesize
\setlength{\tabcolsep}{3pt}
\begin{tabular}{@{}lccccc@{}}
\toprule
Archetype & $f$ & BER aligned & Predicted & Measured & $G$ \\
\midrule
barrage & $1.000$ & $0.480$ & $0.480$ & $0.481$ & $0.0$ \\
fixed-partial & $0.250$ & $0.480$ & $0.400$ & $0.401$ & $6.0$ \\
blind sweep & $0.122$ & $0.484$ & $0.389$ & $0.390$ & $9.1$ \\
\bottomrule
\end{tabular}

\medskip
\parbox{\columnwidth}{\scriptsize $f$: jammed fraction; $G=-10\log_{10}f$: geometric gain over the aligned case; aligned: jammer parked on the carrier; predicted: \eqref{eq:mixing}. The reactive follower is absent because it never engages ($f{=}0$), so its hop BER equals the clean one; the test runs on the TDL-D link with the frozen receivers, whose clean BER is $\approx0.375$.}
\end{table}

\textbf{SWaP-C footprint.}\label{subsec:latency} Table~\ref{tab:swapc} collects the SWaP-C budget from the deployed artifacts rather than from estimates: the float32 parameter count, the measured int8 flatbuffer, the measured activation arena, and the single-burst latency of the full int8 firmware running on a $240$~MHz ESP32-S3 (TensorFlow Lite Micro with the ESP-NN int8 kernels, no accelerator). Sizes are in KiB and the latency in ms. Both receivers are the most accurate in the operating region and the only ones for which a microcontroller deployment is demonstrated. The Conv1D variant is the smallest and fastest overall ($27{,}620$ parameters, $45.6$~KiB of int8 weights, $24.2$~KiB of activations, $16.90$~ms per burst, $59$ bursts/s); the QKV variant pays $1.6\times$ the parameters and $1.7\times$ the int8 flatbuffer, $7.7\times$ the activations and $11.5\times$ the latency for its attention bottleneck, and is therefore the variant for an accelerator-equipped edge node rather than for a flight controller. The baseline receivers are not deployed and are not tabulated, because the comparison would not be like for like --- TensorFlow Lite Micro lowers the recurrent layer to a \texttt{UnidirectionalSequenceLSTM} kernel that the ESP-NN int8 set does not accelerate and that needs persistent state tensors, so a latency measured there would characterize the runtime rather than the architecture. The RF front end is outside this digital accounting: the circulator/duplexer adds insertion loss and an isolation constraint, and the $40$~dB SIC figure is a design budget rather than a measured micro-UAV front end. This makes the $40$~dB choice conservative for the sensing result, because obstacle-ID varies by at most one point from $0$ to $50$~dB in Sec.~\ref{subsec:channel_model}.

\begin{table}[!htb]
\centering
\caption{Deployment cost on the ESP32-S3.}
\label{tab:swapc}
\footnotesize
\setlength{\tabcolsep}{2.2pt}
\begin{tabular}{@{}lccccc@{}}
\toprule
Receiver & params & f32 & int8 & arena & $t_r$ [ms] \\
\midrule
Ultra-CAN (Conv1D) & $27{,}620$ & $107.9$ & $45.6$ & $24.2$ & $16.90$ \\
Ultra-CAN-QKV & $44{,}195$ & $172.6$ & $78.8$ & $185.6$ & $193.66$ \\

\bottomrule
\end{tabular}

\medskip
\parbox{\columnwidth}{\scriptsize sizes in KiB and latency in ms; params: float32 parameter count; f32: float32 weights; int8: quantized flatbuffer; arena: activation arena; $t_r$: latency of one burst.}
\end{table}

\subsection{Deployment on a Low-Cost Microcontroller}
\label{subsec:mcu}

\begin{figure}[!t]
\centering
\includegraphics[width=0.9\columnwidth]{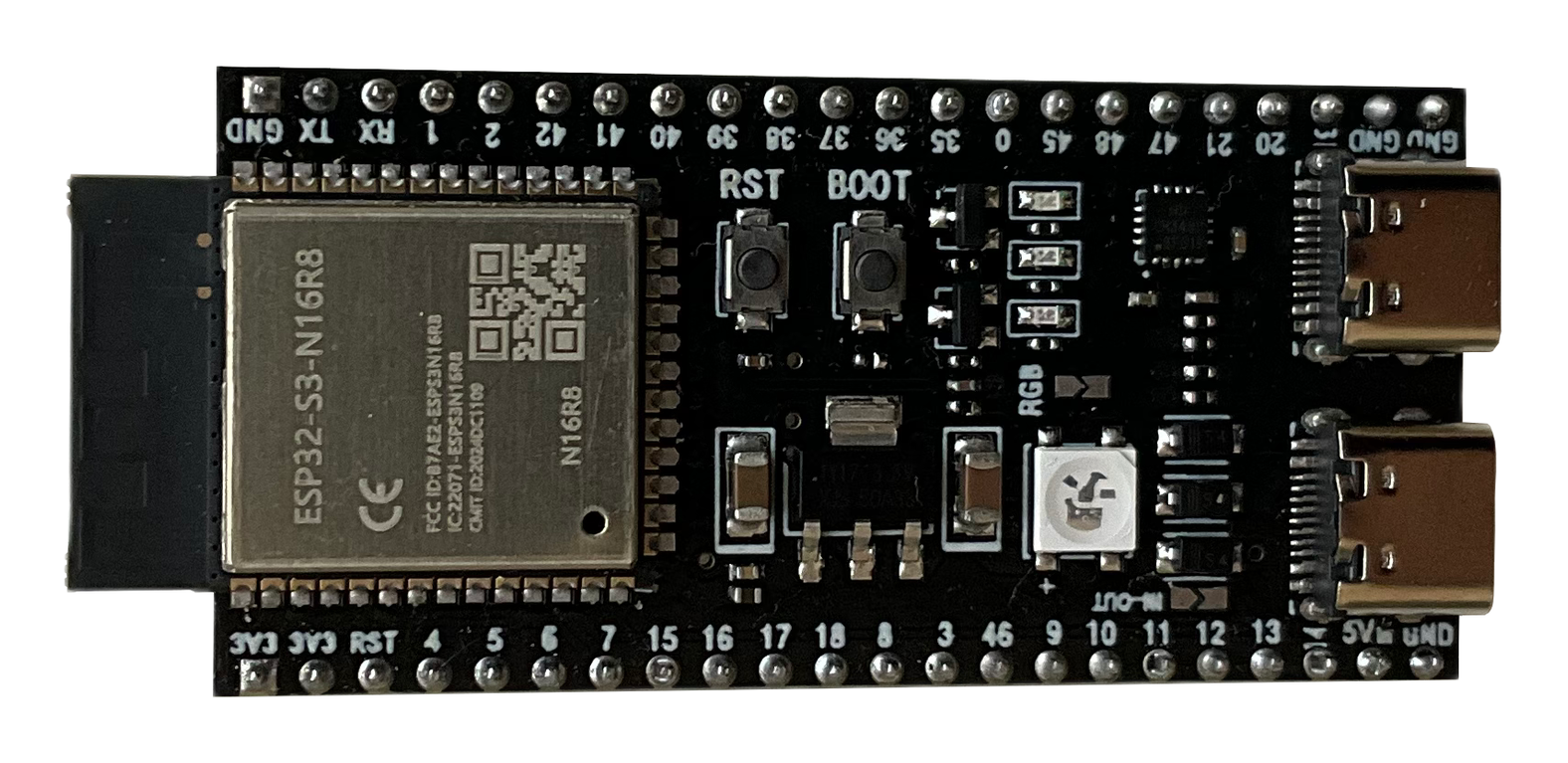}
\caption{Deployment platform: the ESP32-S3-DevKitC-1 ($240$~MHz, $512$~KiB of SRAM, no accelerator) on which both int8 receivers were measured, running TensorFlow Lite Micro with the ESP-NN int8 kernels. The $10$~MS/s I/Q front end that feeds the two input channels is outside this digital accounting.}
\label{fig:esp32}
\end{figure}

The receiver is only useful if it can be embedded, so both variants were exported to int8 TensorFlow Lite Micro firmware and measured on a $240$~MHz ESP32-S3 with $512$~KiB of SRAM and no accelerator (Fig.~\ref{fig:esp32}). Post-training quantization is transparent on both outputs. Against the float model, evaluated on $20{,}000$ bursts per SNR point, the int8 receiver changes at most two decisions in $20{,}000$ (BER $0.00355$ against $0.00350$ at $5$~dB, $0.00085$ against $0.00090$ at $13$~dB) and preserves the sensing outcome: the obstacle-ID rate $66.7\%$ against $66.6\%$ and the peer-report rate $4.6\%$ against $4.3\%$, and at most $0.02\%$ peer reports on the answered bursts at $\gamma{=}1.5$, against the $0.05\%$ bound of Sec.~\ref{subsec:peers_eval}. The gate is quantization-immune by construction, because it thresholds the local-seed delay profile, which is computed outside the network; its answered fraction is therefore identical in both precisions. Quantizing the shipped model with a noise-like representative dataset instead of recorded bursts is \emph{not} transparent (it drives the receiver to chance level), which is worth stating as an implementation constraint: the calibration set must be drawn from the operating channel.

\textbf{Three clocks.} The deployment is described by three rates. The waveform carries one bit per $10\,\mu$s burst, i.e. $100$~kbit/s; a dwell holds the channel for $D$ bursts; and the receiver emits one decision every $t_r$. A receiver that skips bursts can only afford to do so if it samples once per dwell and the channel is constant over the gap, hence
\begin{equation}
D_{\min}=\left\lceil \frac{t_r}{10\,\mu\mathrm{s}}\right\rceil,\;
R_{\mathrm{hop}}=\frac{1}{D_{\min}\cdot10\,\mu\mathrm{s}},\;
R_b=\frac{100~\mathrm{kbit/s}}{D_{\min}}.
\label{eq:rate}
\end{equation}
Here $D_{\min}$ is the dwell that the \emph{platform} imposes: the number of $10\,\mu$s slots that one decision occupies, so that one decoded burst per dwell is the fastest rate the node can sustain. It is a property of the hardware and is not the $D\in\{1,2,4,8,16\}$ that the transmitter selects in the hopping protocol of Sec.~\ref{subsec:threat_model}. Table~\ref{tab:rates} evaluates \eqref{eq:rate} on the measured platforms, where $t_r$ is the median single-burst latency, $R_{\mathrm{hop}}$ the hop rate, $R_b$ the decoded payload and the last column the update rate of a closing-velocity estimate (eight bursts per track, Sec.~\ref{subsec:velocity}). On the microcontroller the Conv1D receiver sustains $59$ bursts/s. Those bursts are shared by the two modes, so the same node delivers up to $59$~bit/s of telemetry, or a $7.4$~Hz closing-velocity estimate when the eight bursts of a track are taken back-to-back ($0.135$~s, Sec.~\ref{subsec:sensing}; at the $100$~ms cadence of Sec.~\ref{subsec:velocity} the same track spans $0.7$~s, i.e. $1.25$~Hz), or any time-multiplexed split of the two. The reference CPU sustains $6.67$~kbit/s; because $D_{\min}=15$ is shorter than the slowest hopping setting of Sec.~\ref{subsec:agility} ($D{=}16$), that setting is decoded at one burst per hop, i.e. at $6.25$~kbit/s of payload. The full $100$~kbit/s air rate requires $t_r\le10\,\mu$s, i.e. $\approx0.1$--$0.3$~TMAC/s: an int8 accelerator, not a flight controller. The per-burst decision quality is the same in all three cases; only the rate changes.

\begin{table}[!htb]
\centering
\caption{Int8 firmware operating points.}
\label{tab:rates}
\footnotesize
\setlength{\tabcolsep}{2.5pt}
\begin{tabular}{@{}lccccc@{}}
\toprule
Platform & $t_r$ [ms] & $D_{\min}$ & $R_{\mathrm{hop}}$ & $R_b$ & track \\
\midrule
ESP32-S3, $240$~MHz & $16.90$ & $1690$ & $59$~Hz & $59$~bit/s & $7.4$~Hz \\
Reference CPU & $0.148$ & $15$ & $6.67$~kHz & $6.67$~kbit/s & $834$~Hz \\
int8 accelerator (est.) & $\approx0.007$ & $1$ & $100$~kHz & $100$~kbit/s & $12.5$~kHz \\
\bottomrule
\end{tabular}

\medskip
\parbox{\columnwidth}{\scriptsize $t_r$: median latency of one decision, in ms; $D_{\min}=\lceil t_r/(10\,\mu\mathrm{s})\rceil$ with $t_r$ in $\mu$s, so $16.90$~ms $=16\,900\,\mu$s gives $D_{\min}=1690$: minimum dwell, in $10\,\mu$s slots; $R_{\mathrm{hop}}$: hop rate; $R_b$: decoded payload; track: update rate of the closing-velocity estimate for eight contiguous bursts; the accuracy of Sec.~\ref{subsec:velocity} is measured at the $100$~ms cadence.}
\end{table}

\textbf{Where the time goes.} The Conv1D receiver spends $34\%$ of the burst in its three convolutions (the two backbone layers plus the $1{\times}1$ attention projection), $14\%$ in the element-wise product and $46\%$ in the two read-out reductions (the max pooling over time, $23\%$, and the attention-weighted sum, $23\%$). The ESP-NN kernels accelerate the first two stages by $63\times$ and $5\times$ with respect to the portable reference kernels, with which the same models take $486$~ms (Conv1D) and $576$~ms (QKV) per burst. The QKV variant spends $66\%$ of the burst in the batched attention product and the int8 softmax, which the ESP-NN kernels do not cover; this, and not the device, is why its disadvantage grows from $3.7\times$ on the CPU to $11.5\times$ on the microcontroller. The energy per burst, estimated from the datasheet active current of the module ($40$--$60$~mA at $3.3$~V, radio off) and the measured time, is $2.2$--$3.4$~mJ for Conv1D and $26$--$38$~mJ for QKV, i.e. $0.15$--$0.20$~W for a node that saturates its receiver. The complete firmware occupies $216$~KiB of static RAM, of which $192$~KiB is the arena provisioned for the QKV model, of which $185.6$~KiB are in use (the Conv1D variant needs $24.2$~KiB), and $457$~KiB of flash.

\textbf{Scope.} No over-the-air test and no self-interference measurement are claimed here: the firmware runs the deployed tensors, and the $10$~MS/s I/Q front end that feeds them is outside this digital accounting, exactly as the circulator is. The quoted $t_r$ is the interpreter call on the two-head graph. The local chaotic reference and the $99$ profile correlations that form the descriptor input are computed before it and were measured on the same board at $14$, $36$ and $494\,\mu$s (median of $20$ runs), $544\,\mu$s in all, i.e. $3.2\%$ of $t_r$. Their $1.3\%$ share of the network's MAC is thus smaller than their share of the time, because they run in float while the network runs on the int8 SIMD kernels. Within that scope the microcontroller is a complete sensing node and a low-rate telemetry node, and the $100$~kbit/s link is an accelerator-class operating point of the same weights and the same decisions.

\section{Discussion}
\label{sec:discussion}

Three design lessons follow from the campaign.

\textbf{Supervision, not capacity, carries the range.} The sensing head wins over the matched filter by $24$ points on the multi-echo + peers scenario and by $24$ points on the peer-free multi-echo link, with a descriptor that contains the matched-filter profile itself. The difference is that the head is trained on the \emph{geometric} obstacle label, so it learns to discount the strongest return when the strongest return is not the target. A correlation-only receiver cannot express that decision, and a larger backbone would not learn it from a peak-based label. This is why the multi-task design is worth its complexity: the sensing function is not a by-product of decoding, it is a second supervised task on the same representation.

\textbf{Safety is a per-burst decision.} The gate trades the answered fraction for correctness and keeps peer reports at or below $0.05\%$ among the answered bursts; the ungated peer-report rate is $4.4\%$. A swarm can therefore choose the operating point that its formation controller needs: at $\gamma{=}1.5$ the receiver answers $23.5\%$ of the bursts at $21$~dB, and three of those answers in four fall on the exact obstacle tap. The remaining silence is not a failure of the estimator; it is the honest answer of a single $10\,\mu$s observation to an ambiguous geometry.

\textbf{The receiver is light in memory, heavy in arithmetic.} Two convolutional layers, one attention stage and two small dense heads occupy $45.6$--$78.8$~KiB as int8 weights and $24.2$--$185.6$~KiB of activations, and they decide a burst in $16.9$--$193.7$~ms on a $240$~MHz microcontroller with no accelerator. The Conv1D variant therefore deploys on a low-cost flight controller and sustains $59$ bursts/s, enough for a $7.4$~Hz closing-velocity estimate or a $59$~bit/s status channel (update rates that align well with the typical reaction times of UAV flight controllers for collision avoidance at swarm speeds of $10$--$30$~m/s); the QKV variant needs an accelerator. The reason is arithmetic rather than memory: the Conv1D receiver spends $\approx1.07$~MMAC per bit, two orders of magnitude more than the $\approx10^{4}$~MAC/bit of a chaotic matched filter, so at a fixed compute budget the achievable bit rate is set by the platform, not by the parameter count. That is also why the full $100$~kbit/s air rate is an accelerator-class operating point of the same weights (Sec.~\ref{subsec:mcu}). The same property makes in-domain retraining affordable at the edge \cite{abdollahpour2025_edge_receiver}: the receiver that sees a new channel can be re-fitted from a local capture without changing the architecture, the heads or the interface.

The limits are physical rather than architectural. A single $10\,\mu$s burst cannot resolve the $160$--$480$~Hz echo Doppler, so velocity is a track-level quantity; the $33$-sample window fixes the unambiguous range at $\approx500$~m; and the gate relies on the swarm's geometric prior for the peer positions, which is the same prior that the formation plan already provides. A cooperative-bistatic extension, M-ary chaos-based modulation for higher throughput, an over-the-air test with a measured self-interference budget, and an accelerator-class deployment of the full bit rate are the natural next steps.

\section{Conclusion}
\label{sec:conclusion}

This paper has shown that a single blind neural receiver driven by a chaos-based waveform can carry secure telemetry and proximity sensing in two time-multiplexed modes, with no separate radar transceiver and no waveform state over the air. Ultra-CAN combines a shared Conv1D backbone, two ultra-light aggregation variants, a communication head and a sensing head that classifies the obstacle lag from the same representation. On the four swarm scenarios the two variants occupy the first two places among the learned receivers in the point estimates, with a pooled BER between $4.6\times10^{-4}$ and $2.4\times10^{-3}$, and the sensing head identifies the labeled obstacle in $66.8\%$ of the peer-laden bursts, against the $48\%$ empirical peak-picker ceiling. A per-burst peer-exclusion gate converts the estimate into a safe range, answering $23.5\%$ of the bursts at $\gamma{=}1.5$ with $0.02\%$ peer reports on the answered set (at most $0.05\%$ across the $\gamma$ sweep) and three of those answers in four on the exact obstacle tap, and a track of eight ranges recovers the closing velocity to $0.70$~m/s. The proposed features survive jamming better than the telemetry does, jamming-aware training restores the link under CW and partial-band attacks, frequency hopping adds a $6$--$9$~dB geometric margin, and the waveform keeps the intercept probability of a channelized detector at $P_d\le0.30$. Both receivers are deployed as int8 firmware: the Conv1D variant decides a burst in $16.9$~ms and $2.2$--$3.4$~mJ on a $240$~MHz microcontroller with $24.2$~KiB of activations, preserving both the demodulation and the sensing decisions of the float model, which yields a measured operating point of up to $59$~bit/s of telemetry or a $7.4$~Hz range track on that node, against $100$~kbit/s for an accelerator-class one. The channel model, the data generator, the receiver implementations and the scripts that reproduce every figure and table of this paper are openly available on GitHub.\footnote{\url{https://github.com/ettorecaruso/ultra-can}}

\section*{Author Contributions}
\textbf{Ettore Caruso:} Software, Investigation, Formal analysis, Methodology, Visualization, Writing - Original Draft, Writing - Review \& Editing. 
\textbf{Marco Siino:} Conceptualization, Resources, Supervision, Validation, Writing - Review \& Editing. 
\textbf{Giacomo Morabito:} Supervision, Validation, Writing - Review \& Editing.

\section*{Conflict of Interest}
The authors declare no conflict of interest.

\IEEEtriggeratref{27}
\bibliographystyle{IEEEtran}
\bibliography{paper}
\end{document}